\documentclass[sigconf, 9pt, nonacm]{acmart} 

\AtBeginDocument{%
  }

\usepackage{balance}
\usepackage{booktabs} 
\renewcommand\footnotetextcopyrightpermission[1]{} %
\usepackage{hyperref}
\usepackage{url}
\usepackage{derivative}
\usepackage{soul}
\usepackage{graphicx}
\usepackage{amsmath}
\usepackage{mathtools}
\usepackage{xspace}
\usepackage{booktabs}
\usepackage{xcolor}
\usepackage{tabularx}
\usepackage{color, colortbl}
\usepackage{diagbox}
\usepackage{multirow}
\usepackage{lipsum}
\usepackage{tabularray}
\usepackage{colortbl}
\usepackage{booktabs}
\usepackage{enumitem}
\usepackage{ragged2e}
\usepackage{soul}

\usepackage[caption=false]{subfig}
\usepackage{parcolumns}
\usepackage{listings}
\usepackage{subfiles}
\usepackage{array}

\setlist{leftmargin=3.5mm}
\definecolor{armygreen}{rgb}{0.29, 0.33, 0.13}
\definecolor{darkgreen}{rgb}{0.0, 0.5, 0.0}
\definecolor{deepskyblue}{rgb}{0.0, 0.75, 1.0}
\definecolor{cvprblue}{rgb}
{0.21,0.49,0.74}
\definecolor{magenta}{rgb}
{1.0, 0.0, 1.0}

\newcommand\newsn[1]{\textcolor{black}{#1}}

\newcommand{\Sys}{\textit{LOCUS}\xspace}
\newcommand{\EA}{InFo\xspace}
\newcommand{\DCI}{LaFS\xspace}
\newcommand{\GR}{GRep\xspace}

\newcommand{\parlabel}[1]{\vspace{0.5em}\noindent\textbf{#1}.}

\begin{document}
\title{LOCUS – LOcalization with Channel Uncertainty and Sporadic Energy}

\author{Subrata Biswas}
\affiliation{%
 \institution{Worcester Polytechnic Institute}
 \city{Worcester}
 \country{USA}}
\email{sbiswas@wpi.edu}

\author{Mohammad Nur Hossain Khan}
\affiliation{%
 \institution{Worcester Polytechnic Institute}
 \city{Worcester}
 \country{USA}}
\email{mkhan@wpi.edu}

\author{Violet Colwell}
\affiliation{%
 \institution{Worcester Polytechnic Institute}
 \city{Worcester}
 \country{USA}}
\email{vcolwell@wpi.edu}

\author{Jack Adiletta}
\affiliation{%
 \institution{Worcester Polytechnic Institute}
 \city{Worcester}
 \country{USA}}
\email{jtadiketta@wpi.edu}

\author{Bashima Islam}
\affiliation{%
 \institution{Worcester Polytechnic Institute}
 \city{Worcester}
 \country{USA}}
\email{bislam@wpi.edu}

\begin{abstract}
Accurate sound source localization (SSL), such as direction-of-arrival (DoA) estimation, relies on consistent multichannel data. However, batteryless systems often suffer from missing data due to the stochastic nature of energy harvesting, degrading localization performance. We propose \Sys, a deep learning framework that recovers corrupted features in such settings. \Sys integrates three modules: (1) Information-Weighted Focus (InFo) to identify corrupted regions, (2) Latent Feature Synthesizer (LaFS) to reconstruct missing features, and (3) Guided Replacement (GRep) to restore data without altering valid inputs. \Sys significantly improves DoA accuracy under missing-channel conditions, achieving up to 36.91\% error reduction on DCASE and LargeSet, and 25.87–59.46\% gains in real-world deployments. We release a 50-hour multichannel dataset to support future research on localization under energy constraints. Our code and data are available at:  \textcolor{magenta}{\url{{https://bashlab.github.io/locus_project/}}}

\end{abstract}

\ccsdesc[500]{Computer systems organization~Sensors and actuators}
\ccsdesc[500]{Computer systems organization~Embedded software}

\keywords{Sound Source Localization (SSL), Feature recovery, Batteryless embedded systems, Deep Neural Network (DNN).}
\maketitle 
\renewcommand{\shortauthors}{Biswas et al.}

\section{Introduction}
\label{sec:intro}

Direction-of-arrival (DoA) estimation enables sound source localization (SSL) by extracting spatial and temporal cues from multichannel microphone data. Classical signal processing methods~\cite{schmidt1986multiple, dibiase2000high} and deep learning approaches~\cite{yalta2017sound, chakrabarty2017multi} have been applied to DoA estimation, with deep learning offering better robustness to real-world challenges like multipath and reverberation.

Intermittently powered batteryless systems face challenges in acquiring full-channel data due to the stochastic nature of energy harvesting. These systems operate using capacitors that store harvested energy, enabling brief active phases followed by sleep when energy depletes.  While capacitor size can control these cycles, longer events often span multiple sleep phases, causing data loss~\cite{monjur2023soundsieve}. Since DoA estimation requires consistent multichannel input, these interruptions severely degrade localization accuracy. As a result, batteryless systems are typically limited to short classification tasks~\cite{9111002, islam2020zygarde, monjur2023soundsieve}, and rarely applied to DoA estimation. Still, their low-power operation holds strong potential for applications like environmental monitoring and wildlife tracking, where source localization is essential.

Existing methods for recovering missing data in audio streams rely on statistical imputation—mean~\cite{allison2001missing}, hot deck~\cite{little2019statistical}, or multiple imputations~\cite{rubin1996multiple}—but suffer from oversimplification, mismatch sensitivity, or high computational cost~\cite{hasan2021missing}. Mutual information-based techniques~\cite{vergara2014review} can detect corrupted regions but fail to recover features effectively for DNN-based DoA estimation.

Recent DNN-based imputation pipelines for batteryless systems    \cite{monjur2023soundsieve} address data recovery but are limited to single-channel input, leaving multichannel recovery for DoA estimation an open challenge. Instead of reconstructing raw audio, prior works suggest that recovering key feature representations—such as generalized cross-correlation (GCC) and Mel-- Frequency Cepstral Coefficients (MFCC) is more power-efficient and preserves spatial localization cues~\cite{de2018paws, gobieski2018intermittent}. This approach is particularly well-suited for batteryless devices, where transmission cost dominates power consumption.

\newsn{Batteryless devices offer a compelling solution for long-term, maintenance-free sensing in challenging or resource-constrained settings. Motivated by the need for spatial awareness in such scenarios, we design our system to perform DoA estimation under the energy and data constraints typical of batteryless platforms, making spatial intelligence feasible where it was previously impractical.} To enable robust DoA estimation in batteryless deployments with intermittent multichannel data loss, we propose \Sys, an offline deep learning framework that reconstructs corrupted multichannel features for sound source localization. Rather than recovering raw audio—which is power-intensive to transmit and process—\Sys focuses on reconstructing key feature representations.


\Sys comprises three modules: (1) Information-Weighted Focus (\EA), which estimates entropy to locate unreliable features; (2) Latent Feature Synthesizer (\DCI), which reconstructs missing values based on learned relationships; and (3) Guided Replacement (\GR), which replaces only corrupted elements while preserving valid data. The reconstructed features are then passed to a downstream DNN for localization. \newsn{While the components used in \Sys are studied separately in previous literature, their integration under the same modular framework enables existing SSL models to function robustly under non-ideal conditions, without requiring changes to SSL model architecture.} Our contributions could be summarized as follows.

\begin{itemize}[noitemsep,topsep=2pt]
\item We propose \Sys, a three-stage framework that recovers corrupted multichannel features using entropy-guided interpolation, enabling robust localization in batteryless systems.
\item We evaluate \Sys across audio and non-audio tasks—on benchmark datasets, and real-world solar and RF-powered environments—and release \textit{LargeSet}, a 50-hour multichannel dataset for future research.
\end{itemize}

While many approaches address sparse data recovery, \Sys is unique in targeting feature-domain recovery under real-world spatial and temporal dropout in multichannel arrays. Unlike compressed sensing or matrix completion methods, \Sys supports partial observability, asynchronous channels, and energy-constrained operation. Its physically inspired attention modules capture spatial cues, which are essential for sound localization capabilities and are not modeled by general-purpose recovery methods.

\Sys achieves up to 21.27\% and 17.14\% lower DoA error than $SoundSeive$ \cite{monjur2023soundsieve} on DCASE \cite{politis2021dataset} and LargeSet, respectively, and improves real-world performance by 25.87–59.46\% under stochastic dropout. It also reduces 90th percentile localization error by 25.52\% in Wi-Fi CSI-based distance estimation \cite{ayyalasomayajula2020deep}. \Sys is well-suited for low-power deployments such as smart buildings and wildlife monitoring, where spatial feature recovery—rather than waveform reconstruction—is sufficient for accurate localization.

\section{Missing Data in Batteryless Microphone Arrays}

\parlabel{Missing Channels in Intermittent Systems}
Batteryless systems powered by RF, solar, or vibration harvesters often experience unpredictable power fluctuations, leading to data loss in multi-channel microphone arrays. Environmental dynamics—like human movement, light occlusion, or mechanical interference—cause inconsistent power, resulting in intermittent microphone failures. In distributed setups, spatial variations in harvested energy lead to asynchronous channel availability, while monolithic systems may deactivate subsets of microphones to conserve power. Although full dropout is possible, partial and asynchronous channel loss is far more common and presents a recurring challenge for localization. As shown in Section~\ref{real_world_eval}, these power variations significantly degrade localization accuracy. \Sys addresses this by reconstructing features and estimating source locations offline, making it ideal for batteryless and low-power systems without requiring on-device inference.

\parlabel{Quantifying Missing Data (MDP)}
We define a channel as the full signal from a single microphone in a microphone array. A channel is considered missing if its entire data stream is absent due to energy loss or simulated dropout. To quantify missingness, we define the Missing Data Percentage (MDP) as \newsn{a temporal metric}, where the fraction of time during which one or more channels are unavailable:
\[
MDP = \frac{t_1 + t_2 + \cdots + t_n}{T} \times 100
\]
where \( t_1, t_2, \ldots, t_n \) are the time segments with one or more missing channels, and \( T \) is the total audio duration.

\begin{figure}[tb]
\centering
\includegraphics[width=0.4\textwidth]{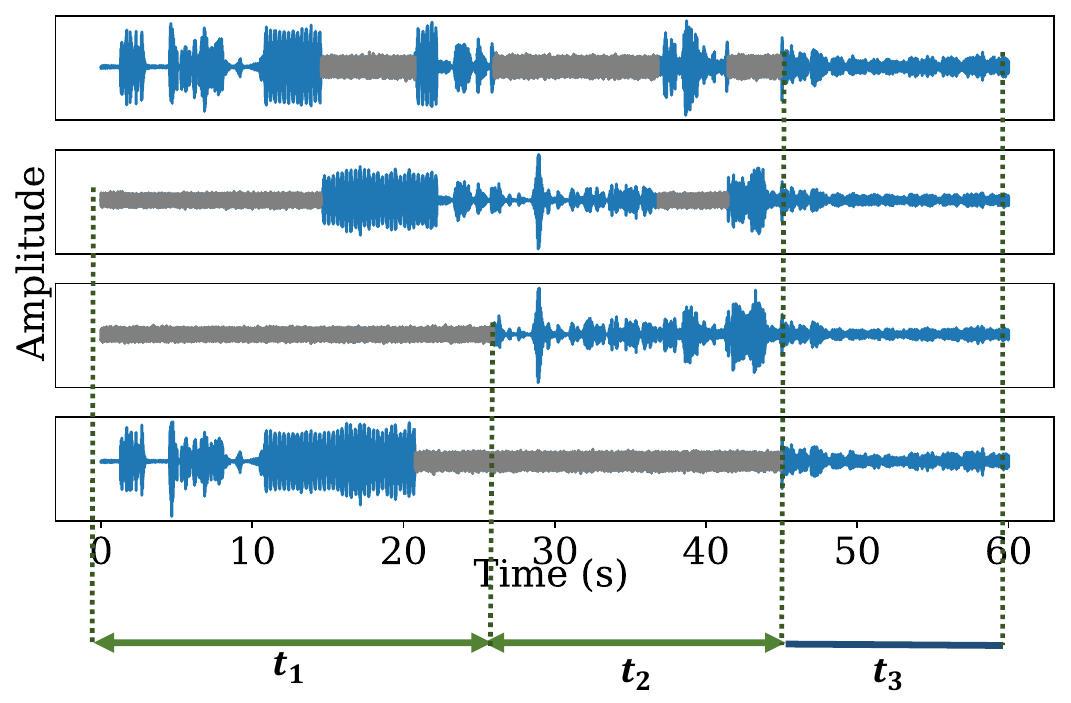}
\caption{Example of raw audio from a four-channel microphone array. Each row represents one microphone; gray regions indicate missing data (75\% MDP).}
\label{exmple_missing_raw_data}
\end{figure}

Figure~\ref{exmple_missing_raw_data} shows an example with 75\% MDP: during segments $t_1$ and $t_2$, at least one channel is missing, while $t_3$ has full data across all microphones. Since 75\% of the total duration includes missing channels, the MDP is 75\%. \newsn{The temporal MDP is preferred over a channel-count-based MDP because it aligns with realistic, intermittent dropout patterns encountered in deployment, capturing both timing and duration of corruption, and enabling consistent evaluation across synthetic and real scenarios.}

\begin{figure}[tb]
\centering
\subfloat[Without missing channels.\label{actual_example}]{\includegraphics[width=.24\textwidth]{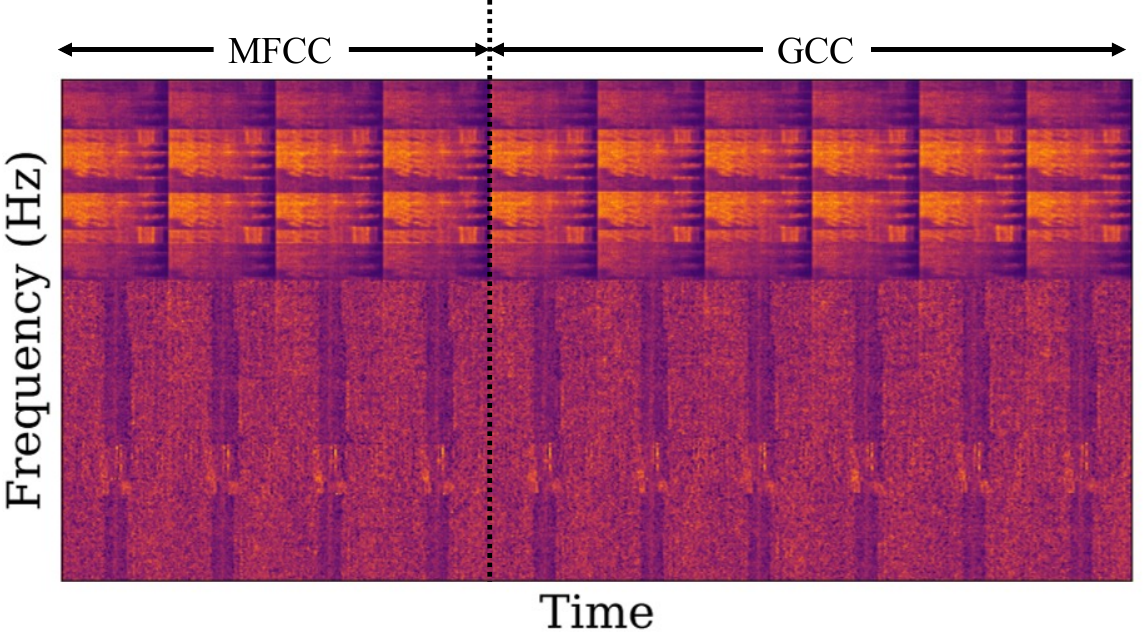}}
\subfloat[With missing channels. \label{doped_example}]{\includegraphics[width=.24\textwidth]{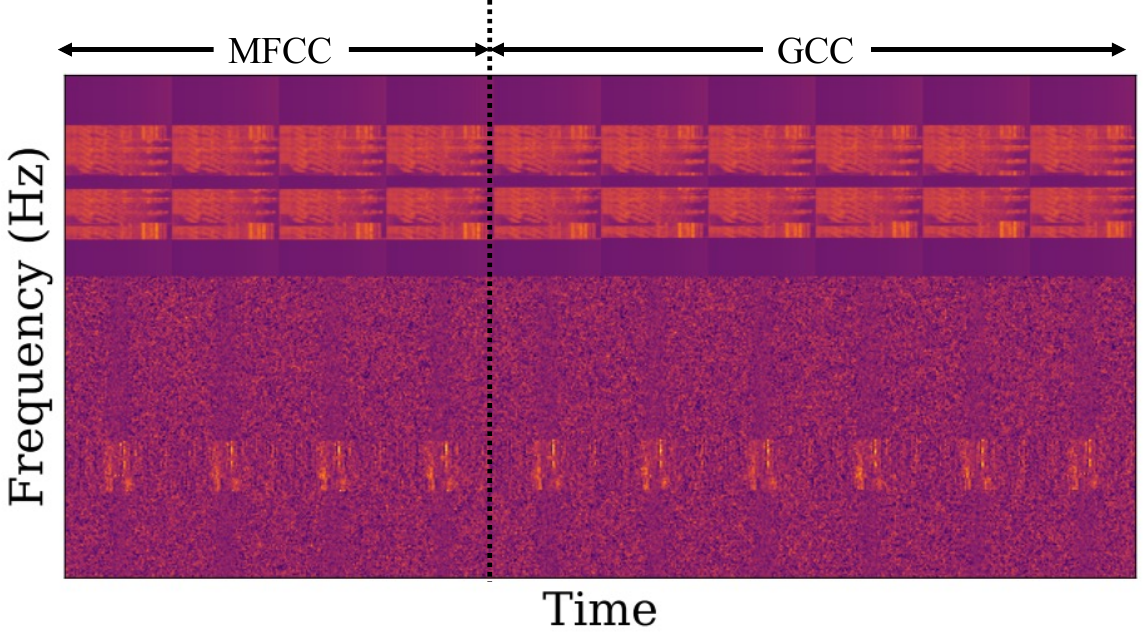}}
\caption{Comparison of multi-channel features with and without missing data. Left: Clean MFCC and GCC; Right: Corrupted versions with missing channels.}
\label{actual_dopped_mel_example}
\end{figure}

\begin{figure}[t]
\centering
\includegraphics[width=0.4\textwidth]{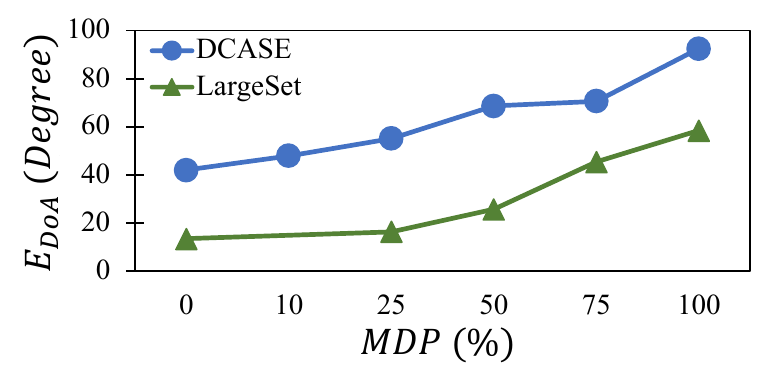}
\caption{DoA error ($E_{DoA}$) increases as the Missing Data Percentage (MDP) rises. Results shown for a baseline SSL model evaluated on two datasets (details in Section~\ref{sec_datset}) under varying levels of missing data.}
\label{fig:mdp_effect}
\end{figure}

\parlabel{Impact of Missing Channels on SSL}
While SSL has been widely studied in both simulated and real-world settings~\cite{evers2020locata,tan2021sound}, the impact of missing channel data remains underexplored. Key features like Mel-frequency cepstral coefficients (MFCC) and Generalized Cross-Correlation (GCC) are highly sensitive to missing inputs, distorting the feature space and degrading DNN performance. As shown in Figure~\ref{actual_dopped_mel_example}, missing channels significantly alter MFCC and GCC patterns. We quantify this by evaluating a standard SSL model~\cite{shimada2021accdoa, adavanne2018sound} on two datasets under increasing MDP levels; DoA error rises sharply (Figure~\ref{fig:mdp_effect}), underscoring the need for more robust recovery methods.

\begin{figure}[t]
\centering
\includegraphics[width=0.45\textwidth]{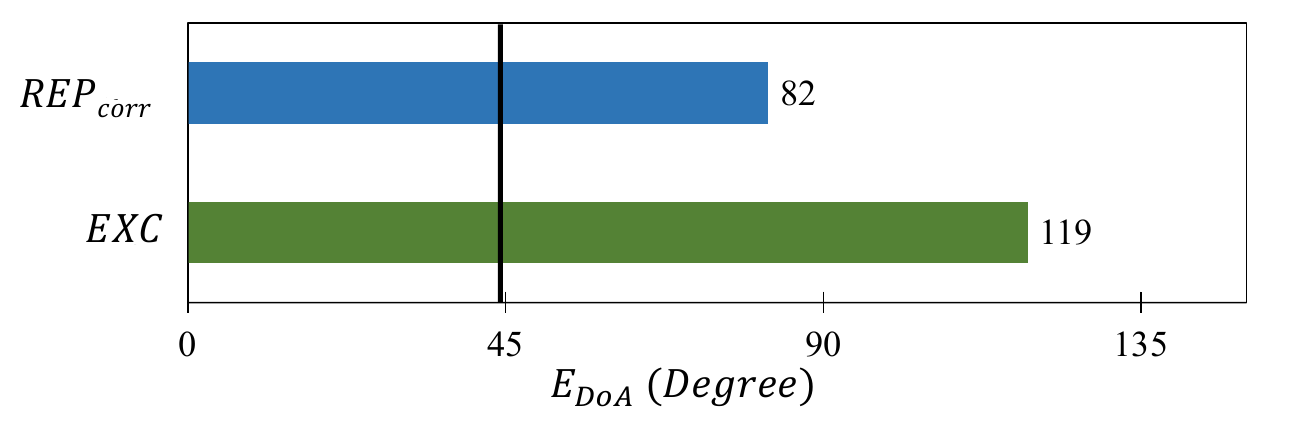}
\caption{DoA error increases when using naive recovery methods: replacing with the most correlated channel ($REP_{corr}$) or excluding faulty channels ($EXC$).}
\label{why_infonet}
\end{figure}

\parlabel{Challenges in Replacing Corrupted Channels}
Although nearby microphones are often correlated, naive approaches such as channel exclusion or correlated-channel replacement fail under dynamic dropout. We tested two such methods at 75\% MDP: $REP_{corr}$, which replaces missing data with the most correlated working microphone’s signal; and $EXC$, which excludes missing channels and retrains the DNN on only the active ones. As shown in Figure~\ref{why_infonet}, both approaches increased DoA error by 46–63\%. These results highlight the need for adaptive strategies like entropy-based recovery, which operate without prior knowledge of which channels are corrupted.

\section{Problem Formulation and Mathematical Foundation}
\label{sec:problem}

\newsn{We begin by revisiting the concepts of \textbf{Full-Rank and Low-Rank Matrices}. A full-rank matrix \( F \) consists of linearly independent rows and columns, implying that all channels contribute distinct and complete information—critical for accurate sound source localization. In contrast, a low-rank matrix \( \tilde{F} \) contains linearly dependent rows or columns, typically arising from missing or corrupted channel data, thereby degrading localization performance.}

\newsn{Let \( X^{n \times T} \) represent the raw multichannel time-domain signal, where \( n \) denotes the number of channels and \( T \) is the time duration per channel. We define \( \mathcal{X} = \{x_1, x_2, \ldots, x_n\} \) as the set of input signals. A feature extraction function \( G \) maps this input to a feature domain as \( F^{n \times D} = G(X) \), where \( D \) is the feature length per channel.}

\newsn{When all channels are present and functioning, \( F \) is full-rank. If one or more channels are missing or degraded, we denote the corrupted input as \( \tilde{X}^{n' \times T} \) with \( n' < n \), and its corresponding features as \( \tilde{F}^{n' \times D} = G(\tilde{X}) \). Our objective is to recover an estimate \( \hat{F} \) that closely approximates the full-rank feature matrix \( F \), i.e., \( \hat{F} \approx F \).}

\newsn{To achieve this, we initially interpolate the missing features from \( \tilde{F} \), yielding \( \bar{\bar{F}} \). However, this interpolation can inadvertently distort the original information, introducing additional error. To mitigate this, we reintroduce the relatively more reliable parts of \( \tilde{F} \) as feedback into the interpolated features.}

\newsn{Because the reliability of individual elements in \( \tilde{F} \) is unknown, we introduce a heuristic confidence mask \( I(\tilde{F}) \), which estimates the presence or absence of information. This mask is computed for each element \( (j,k) \) of \( \tilde{F} \) as:}
\begin{equation}
I_{j, k}(\tilde{F}) = \sigma(- \tilde{F}_{j, k} \log \tilde{F}_{j, k})
\label{eq_i}
\end{equation}

\newsn{Here, \( \sigma \) is the sigmoid function that bounds the confidence values between 0 and 1. This formulation is not derived from a formal probabilistic model, but rather serves as a heuristic measure of reliability. A value of 1 indicates high confidence in the information content of \( \tilde{F}_{j,k} \), while 0 indicates complete absence, with intermediate values suggesting partial reliability.}

\newsn{Using this confidence mask, we refine the interpolated feature map as:}
\begin{equation}
\hat{F} = \bar{\bar{F}} + I(\tilde{F}) \otimes (\bar{\bar{F}} - \tilde{F})
\label{eqn:feedback}
\end{equation}

\newsn{This expression can be equivalently rewritten as:}
\begin{equation}
\hat{F} = I(\tilde{F}) \otimes \tilde{F} + (1 - I(\tilde{F})) \otimes \bar{\bar{F}}
\label{replacement_final}
\end{equation}

\newsn{In this formulation, \( I(\tilde{F}) \) acts as a confidence-driven selector: it retains elements from \( \tilde{F} \) where the confidence is high and replaces uncertain elements with interpolated values from \( \bar{\bar{F}} \). Our goal throughout this paper is to estimate a full-rank feature matrix \( \hat{F} \) from a corrupted, low-rank feature matrix \( \tilde{F} \) using this reliability-guided reconstruction process.}

\section{\Sys: LOcalization with Channel Uncertainty and Sporadic Energy}

\begin{figure*}[t]
\centering
   \includegraphics[width=\textwidth]{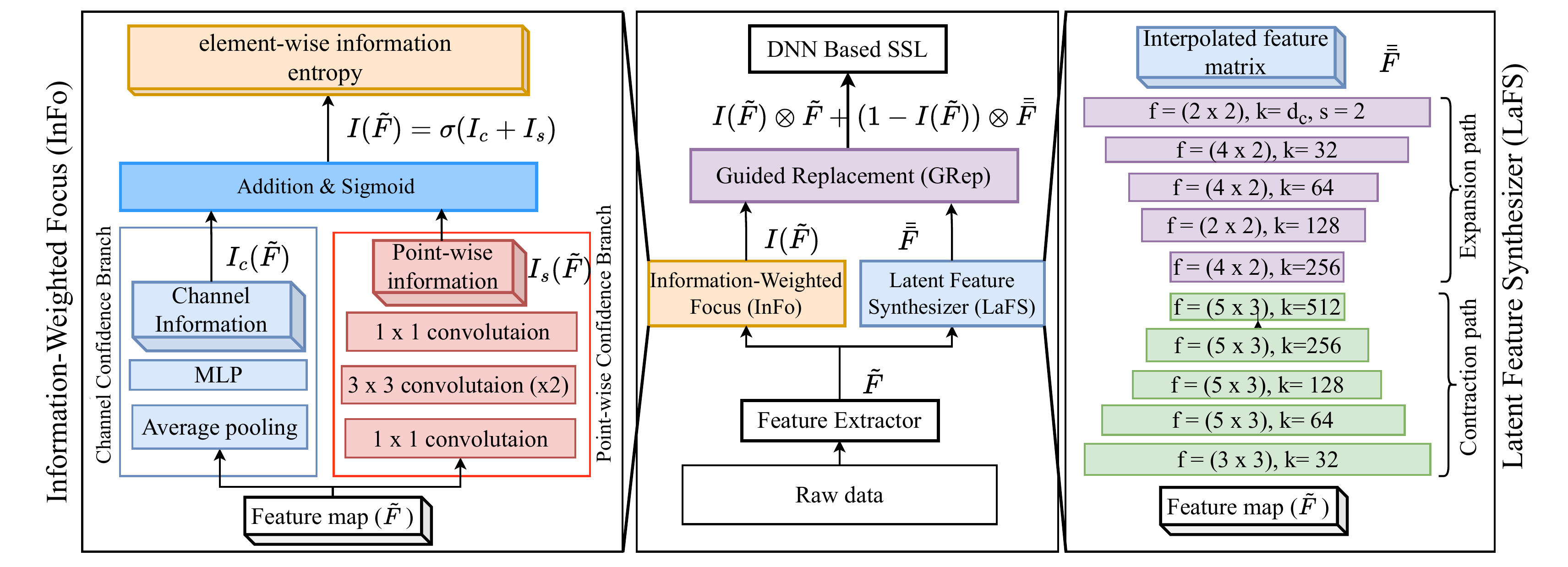}
  \caption{\Sys consists of three key components for robust feature recovery under missing data. InFo estimates element-wise \newsn{confidence score}, $I(\tilde{F})$, from the low-rank feature map $\tilde{F}$ using both channel-wise and spatial feature branches. LaFS reconstructs a full-rank feature representation $\bar{\bar{F}}$ by modeling dependencies across available channels via a symmetric encoder-decoder structure. GRep combines the original and synthesized features using $I(\tilde{F})$ to produce the final recovered feature map $\hat{F}$.}
  \label{model_arch}
\end{figure*}

\Sys estimates the full-rank feature matrix $\hat{F} \approx F$ from a corrupted low-rank matrix $\tilde{F}$, following the recovery formulation in Equation~\ref{replacement_final}. As shown in Figure~\ref{model_arch}, \Sys operates between the feature extractor and the DNN-based sound source localization (SSL) model. It consists of three key components:
(1) \newsn{Information-Weighted Focus (\EA), which estimates element-wise confidence score $I(\tilde{F})$;}
(2) \newsn{Latent Feature Synthesizer (\DCI), which reconstructs a complete feature set $\bar{\bar{F}}$ from $\tilde{F}$;} and
(3) \newsn{Guided Replacement (\GR), which fuses $\tilde{F}$ and $\bar{\bar{F}}$ using confidence masking $I(\tilde{F})$ to produce $\hat{F}$.
This process is governed by Equation~\ref{replacement_final}, and the resulting $\hat{F}$ is passed to the downstream SSL model for inference.}

\subsection{InFo: \underline{In}formation-Weighted \underline{Fo}cus}

\newsn{The purpose of \textbf{In}formation-Weighted \textbf{Fo}cus (\textbf{\EA}) module is to estimate the confidence score, $I(\tilde{F})$, for each element of the low-rank feature set $\tilde{F}$. The \EA module estimates the channel and spatial confidence score ~\cite{park2018bam} by giving higher attention to elements with more available information. The input to \EA is the low-rank feature set $\tilde{F} \in \mathbb{R}^{d_{\tilde{X}}}$, where $d_{\tilde{X}}$ is the dimension of $\tilde{F}$. The left-hand side of Figure~\ref{model_arch} shows the proposed early attention architecture, consisting of two main components.}

\begin{figure*}[!htb]
    \centering
    \subfloat[Retrieval of available information matrix from low-rank feature $\tilde{F}$. Here, the yellow segment signifies the missing information content in $\tilde{F}$.\label{ea_fig}]{\includegraphics[width=.27\textwidth]{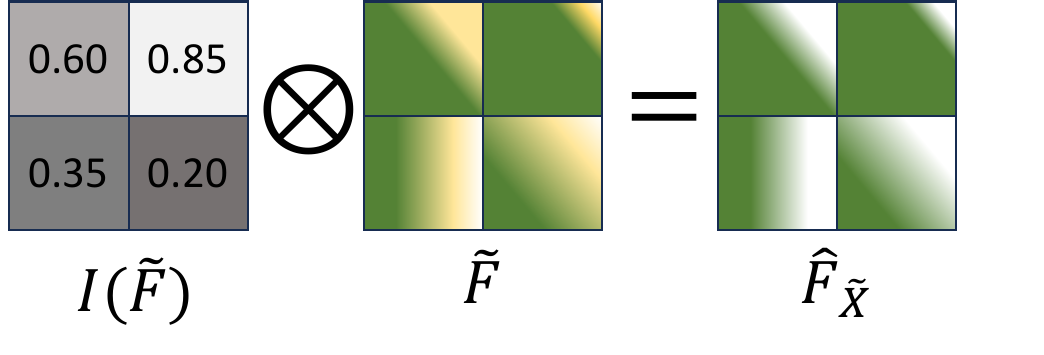}}
    \hspace{.1in}
    \centering
    \subfloat[Retrieval of available $\hat{F}_{Z|\tilde{X}}$ from interpolated feature $\bar{\bar{F}}$. Here, the dotted green sections signify the region where error is introduced in the process of interpolation.\label{dci_fig}] {\includegraphics[width=.37\textwidth]{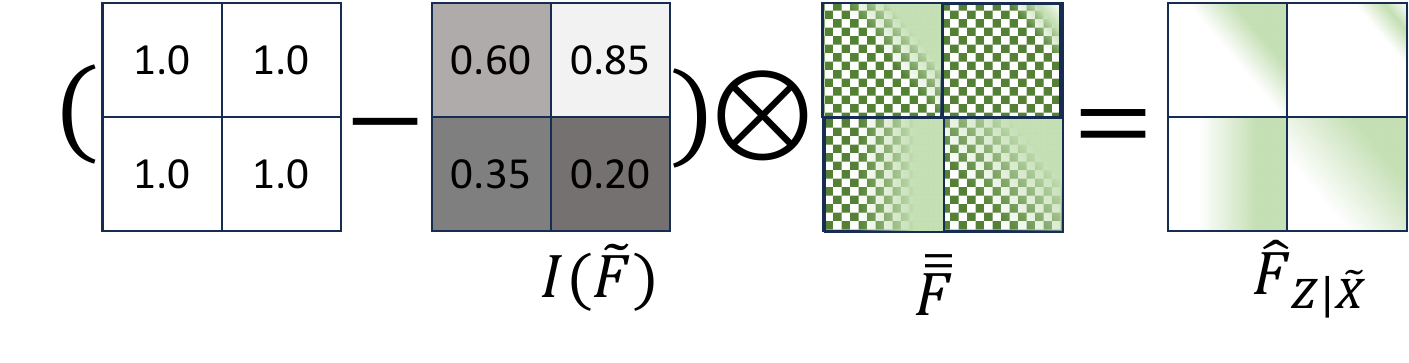}}
    \hspace{.1in}
    \subfloat[Estimating full-rank feature $\hat{F}$ with $\hat{F}_{\tilde{X}}$ and $\hat{F}_{Z|\tilde{X}}$. \label{gr_fig}]{\includegraphics[width=.27\textwidth]{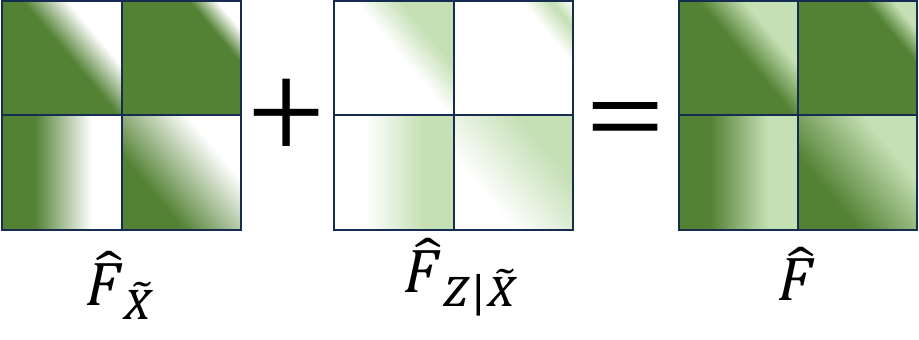}}
    \caption{Step-by-step workflow of Guided Replacement (\GR) to estimate the full-rank feature $\hat{F}$ from the low-rank feature $\tilde{F}$. \newsn{Here confidence score $I(\tilde{F})$ acting as weight. When $I_{j,k}$ indicates low confidence in the original $F_{j,k}$, the system relies more on $\hat{F}{j,k}$; conversely, when $F_{j,k}$ appears reliable, it is preserved.
}}
\end{figure*}

\newsn{The first component of \EA is the \textit{channel confidence branch}, $f_{\gamma}(\tilde{F})$, parameterized with $\gamma$. This branch performs average pooling followed by a multi-layer perceptron to extract the channel confidence score. Average pooling aggregates information across all indices, enabling the model to learn more robust features while accurately depicting the overall strength of a feature. The multi-layer perceptron consists of three fully connected layers with batch normalization and rectified linear unit (ReLU) activation after the first two layers. This branch outputs a vector $I_c = f_{\gamma}(\tilde{F})$ that measures the global confidence score for each channel.}

\newsn{The second component of \EA is the \textit{point-wise confidence branch}, $f_{\lambda}(\tilde{F})$, parameterized with $\lambda$. The input $\tilde{F}$ first passes through a convolution layer with a kernel size of $(1 \times 1)$ and a channel reduction factor of $r$. It then goes through $n$ dilated convolution layers with a kernel size of $(3 \times 3)$, dilation value $d$, batch normalization, and ReLU activation. In our architecture, $n = 3$ and $d = 2$. The dilated convolution increases the receptive field, enabling us to leverage contextual information. Finally, this contextual information passes through another convolutional layer with a kernel size of $(1 \times 1)$ and outputs $I_s = f_{\lambda}(\tilde{F})$, which estimates the point-wise confidence score of the input feature.}

\newsn{Next, we broadcast $I_s$ to \(\mathbb{R}^{d_{\tilde{X}}}\) to match the dimension of $\tilde{F}$ and perform element-wise addition on $I_c$ and $I_s$ to estimate the information entropy. Finally, the \textit{sigmoid} function, $\sigma$, maps the results between 0 and 1, where a value closer to 1 indicates higher information availability and vice versa. Therefore, the confidence score is defined as:}
\begin{equation}
    I(\tilde{F}) = \sigma(I_c + I_s)
\end{equation}
\subsection{LaFS: \underline{La}tent \underline{F}eature \underline{S}ynthesizer}

The \textbf{La}tent \textbf{F}eature \textbf{S}ynthesizer (\textbf{\DCI}) estimates a full-rank feature set, $\bar{\bar{F}}$, from the available low-rank feature set $\tilde{F}$ by exploiting relations among the available channels. The right-hand side of Figure~\ref{model_arch} shows the architecture of \DCI, which takes $\tilde{F}$ as input and provides the interpolated feature $\bar{\bar{F}}$ as output. We choose an auto-encoder for estimating $\bar{\bar{F}}$ as it can learn compressed intermediate representations. Auto-encoders can efficiently interpolate missing information by semantically mixing characteristics from data~\cite{berthelot2018understanding} and are used in latent variable generation~\cite{dumoulin2016adversarially, ha2017neural}.

\DCI contains two paths: (1) contraction and (2) expansion. The contraction path has $5$ downsampling blocks, each consisting of a convolution layer (kernel size = $5 \times 3$), ReLU activation, and batch normalization. The output of the contraction path is the latent intermediate representation $F_{con} \in \mathbb{R}^{d_c}$, where $d_c$ represents the intermediate feature dimension.

The expansion path passes the input $F_{con}$ through 5 consecutive upsampling blocks, each consisting of transposed convolution with batch normalization and leaky-ReLU activation. The final convolution uses sigmoid activation without batch normalization. After each upsampling block, we halve the number of output feature filters, setting the last layer’s output filter number to match the shape of the input feature map. The expansion path returns the interpolated feature matrix, $\bar{\bar{F}} = f_{\phi}(\tilde{F})$, where $\bar{\bar{F}} \in \mathbb{R}^{d_z}$.

\subsection{GRep: \underline{G}uided \underline{Rep}lacement}

\newsn{The \textbf{G}uided \textbf{Rep}lacement (\textbf{\GR}) module recovers the full-rank feature $\hat{F}$ using Equation~\ref{replacement_final}. Figure \ref{ea_fig} depicts this retrieval process. First, we perform element-wise multiplication between the confidence mask $I(\tilde{F})$ and the low-rank feature set $\tilde{F}$ to retrieve the available information in the low-rank feature set $\hat{F}_{\tilde{X}}$.}

\newsn{Next, we find the missing information matrix $\hat{F}_{Z|\tilde{X}}$ from the interpolated feature $\bar{\bar{F}}$. Here, $Z$ represents the missing data in $\tilde{\mathcal{X}}$. As shown in Figure~\ref{dci_fig}, we take the complement of the confidence mask $I(\tilde{F})$ and multiply it element-wise with $\bar{\bar{F}}$ to find the missing information, disregarding regions (dotted darker green) where errors might have been introduced during interpolation.}

Finally, \GR utilizes the available information matrix $\hat{F}_{\tilde{X}}$ and the missing information matrix $\hat{F}_{Z|\tilde{X}}$ to estimate the full-rank feature $\hat{F}$ from Equation~\ref{replacement_final} (Figure~\ref{gr_fig}).
\definecolor{Mercury}{rgb}{0.882,0.882,0.882}
\begin{table}[t]
\centering
\caption{Dataset Description.}
\label{dataset_desc_tab}
\resizebox{0.48\textwidth}{!}{
\begin{tabular}{ccccc}
\toprule[1.5pt]
\textbf{Dataset } & \textbf{Duration} & \textbf{$F_s$(Hz)} & \textbf{No of samples} & \textbf{Train-Val-Test} \\ \midrule[1pt]
DCASE    & 6 Hour   & 24000         & 600           & 400-100-100           \\
LargeSet & 50 hour  & 44100         & 30000         & 21000-3000-6000      \\ \bottomrule[1.5pt] 
\end{tabular}
}
\vspace{-1em}
\end{table}

\section{Experimental Setup and Implementation Details}

This section outlines the datasets, data preparation, implementation, baselines, and evaluation metrics used to assess \Sys. Our codebase is available at \textcolor{magenta}{\url{https://github.com/BASHLab/LOCUS}}.

\subsection{Dataset}
\label{sec_datset}
We evaluate \Sys using two distinct multichannel sound source localization datasets, summarized in Table~\ref{dataset_desc_tab}. \newsn{Our evaluation covers both temporal (DCASE dataset, moving sources) and snapshot-based (LargSet, static sources) DoA tasks, demonstrating \Sys's generalizability across varied temporal reasoning requirements.}

\noindent \textbf{Dataset 1: DCASE.}
The DCASE2021 Task 4 dataset~\cite{politis2021dataset} consists of 600 one-minute polyphonic recordings at 24 kHz, captured with a four-channel microphone array. The dataset includes up to four simultaneous sound sources and 12 distinct acoustic events, with Direction of Arrival (DoA) annotations for each source. It also features directional interference and multi-channel ambient noise, with SNR levels ranging from 6 to 30 dB. The dataset is split into 400 training and 200 evaluation samples.

\begin{figure*}[!htb]
\begin{minipage}{0.62\textwidth}
\centering
    \vspace{-1.5em}\includegraphics[width=\textwidth]{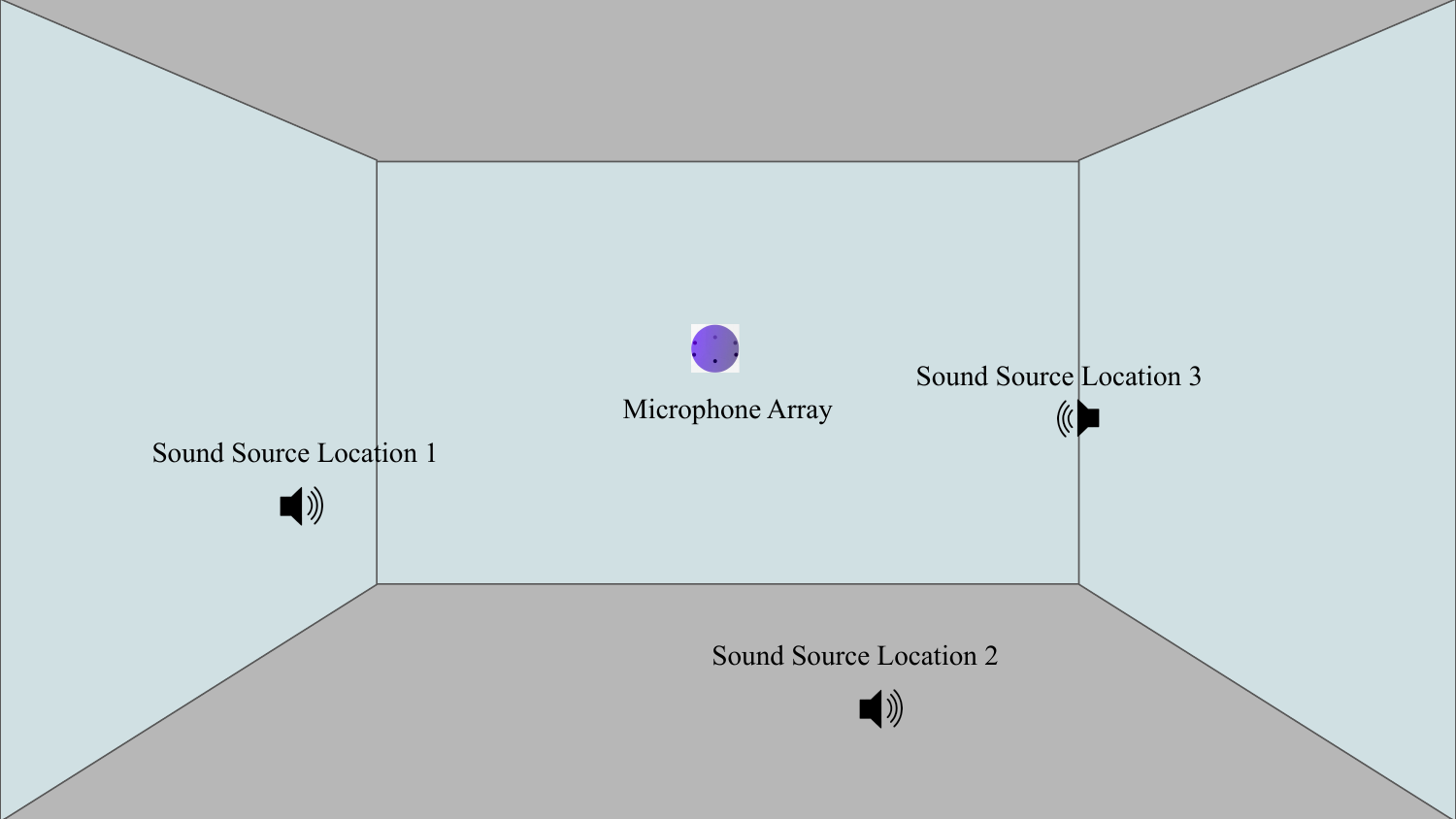}
    \caption{\newsn{Simulation environment with the microphone array in the center.}}
    \label{sim_env}
\end{minipage}
\hspace{0.7em}
\begin{minipage}{0.34\textwidth}
\centering
\vspace{-.5em}
\includegraphics[width=\textwidth]{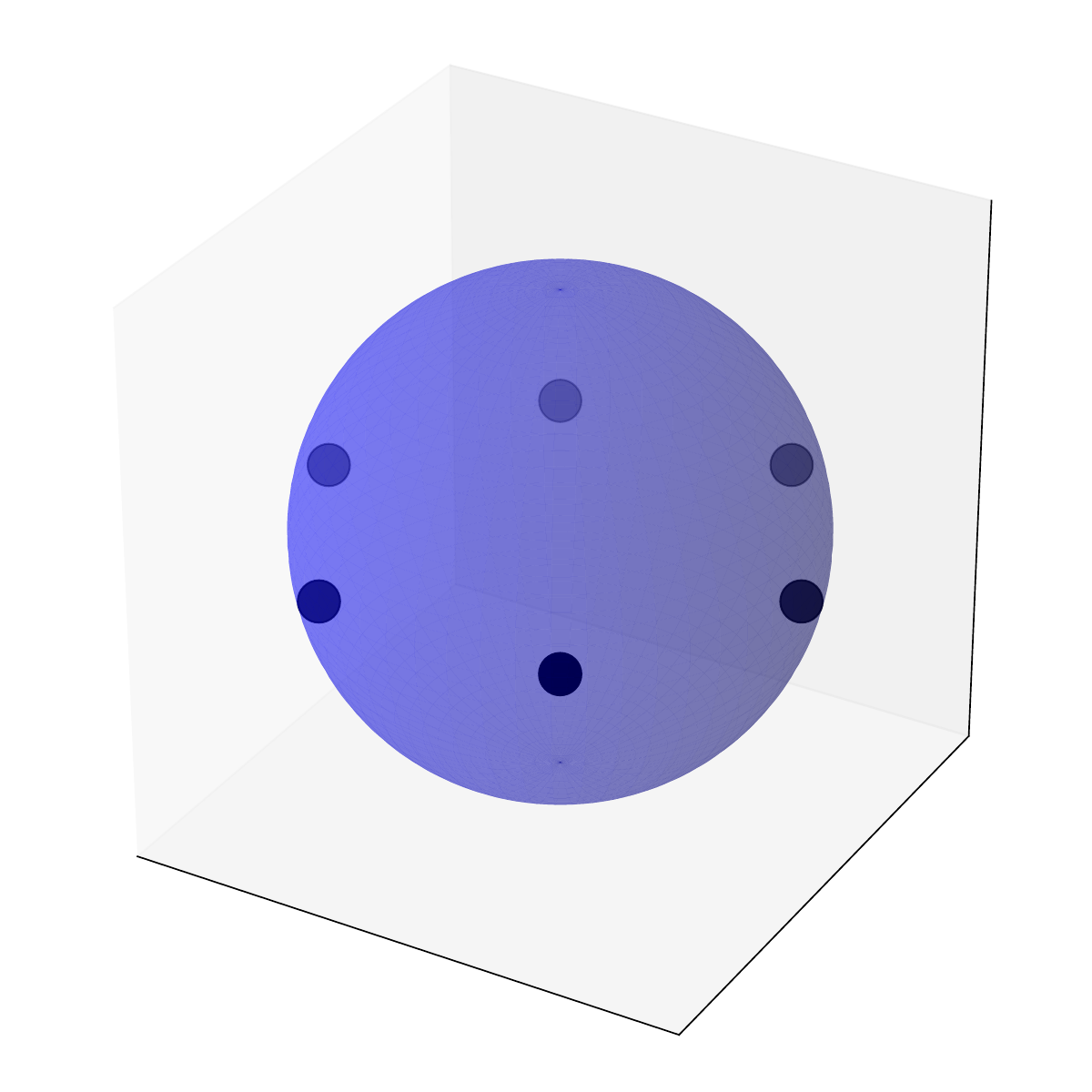}
    \caption{\newsn{Microphone array}}
    \label{mic_array}
\end{minipage}
\vspace{-1em}
\end{figure*}

\begin{figure}[]
\begin{minipage}{0.47\textwidth}
\centering
\captionsetup{type=table}
\label{tab:sim_env}
\definecolor{Mercury}{rgb}{0.882,0.882,0.882}
\caption{LargeSet Simulation Environment}
\label{sim_env_desc}
\resizebox{\textwidth}{!}{
\footnotesize{
\begin{tabular}{
c|c
}
\toprule[1.5pt]
\textbf{Dimension}                 & \textbf{Description} \\ \midrule[1pt]
$10 \times 7.5 \times 3.5$ & unpainted concrete - carpet, open cell foam - reverb chamber \\
$10 \times 7.5 \times 3.5$ & plasterboard - cocos fibre roll - ceramic tiles \\
$15 \times 8 \times 5$     & wooden lining - linoleum on concrete - lime wash \\
$15 \times 8 \times 5$     & hard surface - carpet rubber - brickwork \\
$10 \times 7.5 \times 3.5$ & plasterboard - carpet hairy - wooden lining \\
$10 \times 7.5 \times 3.5$ & lime wash - carpet, closed cell foam - plasterboard \\
$15 \times 8 \times 5$     & lime wash - felt - brick wall \\
$15 \times 8 \times 5$     & lime wash - carpet cotton - concrete \\
$7.5 \times 4 \times 3$    & plasterboard - carpet thin - rough concrete \\
$7.5 \times 4 \times 3$    & concrete - carpet tufted - hard surface \\ \bottomrule[1.5pt]
\end{tabular}
}
}

\vspace{-1.5em}
\end{minipage}
\end{figure}

\noindent \textbf{Dataset 2: LargeSet.}
\label{large_set}
We created a 50-hour, six-channel dataset with 180,000 one-second recordings at 44.1 kHz across 10 diverse environments. This monophonic dataset avoids overlapping sound sources to isolate the impact of missing channels.

\noindent \textbf{Simulation Details.}
We simulate a total of ten distinct acoustic environments \newsn{with  \emph{PyRoomacoustics} \cite{scheibler2018pyroomacoustics} }using the configuration parameters outlined in Table 2. Each environment represents a closed rectangular room with rigid boundaries, comprising a ceiling, floor, and four walls, where the material properties and layout may vary across environments to introduce diversity in reverberation and reflection characteristics. The dimensions of each room are defined in meters as $(l \times w \times h)$, representing the length, width, and height, respectively. \newsn{Figure \ref{sim_env} and figure \ref{mic_array} depict the schematic of the simulation room and the microphone array geometry.}

At the geometric center of each room, located at coordinate $\left(\frac{l}{2}, \frac{w}{2}, \frac{h}{2}\right)$, a spherical microphone array is placed. The array consists of six omnidirectional microphones uniformly distributed along the equator of the sphere. The radius of the microphone array is set to $4.25$ cm, ensuring a compact configuration suitable for near-field sound localization.

\newsn{To simulate sound sources within each environment, we use the \emph{UrbanSound8K} dataset~\cite{salamon2014dataset}, to expand
the diversity of acoustic content (e.g., engines, alarms, dog barks) and evaluate the generalization of our system
beyond typical indoor scenes.. For each simulation instance, a single sound source location is randomly sampled from a continuous uniform distribution within the room volume. Specifically, the source coordinate $(x, y, z)$ is drawn such that $x \in [0, l)$, $y \in [0, w)$, and $z \in [0, h)$, ensuring that the source is located strictly within the room boundaries. This coordinate is recorded as the ground truth for supervised learning tasks such as source localization or direction-of-arrival (DoA) estimation.}


\subsection{Data Preparation}
\noindent\textbf{Feature Extraction.}
\label{feature_extraction}
We extract two key acoustic features: Mel-Frequency Cepstral Coefficients (MFCC) and Generalized Cross-Correlation (GCC) from raw audio using a 1024-point FFT with a 40 ms window and 20 ms hop length at 24 kHz and 44.1 kHz. MFCC captures magnitude-based audio information, while GCC captures time differences and phase shifts between channels, crucial for spatial localization. These features are combined into a matrix, $\tilde{F}$, which is input to the DNN-SSL pipeline.


\noindent\textbf{Missing Channel Perturbation}
\label{data_dopping}
To simulate missing channels, we randomly select $m \leq \frac{n}{2}$ channels from $n$ microphones, where $m$ represents the missing channels. For each Missing Data Percentage (MDP), we replace $p\%$ of the audio with Gaussian noise (mean = 0, std = 1). We also simulate channel failures by randomly dropping entire channels for specific timeframes (100\% MDP). These perturbations enable the model to handle real-world scenarios of microphone failure or signal distortion.


\subsection{Implementation Details}
\noindent\textbf{Training Pipeline of \Sys.}
We train \Sys jointly with the DNN-based sound source localization (SSL) task. The SSL loss function, $\mathcal{L}_{DT}$, backpropagates through the network and the information branches ($f_{\gamma}$ and $f_{\lambda}$) of \EA, enriching the features used for SSL.

The \DCI branch ($f_{\phi}$) has a separate training objective to estimate the input features. Its loss function, $\mathcal{L}_{DCI}$, is the squared $L_2$ distance between the original and reconstructed feature matrices:
$\mathcal{L}_{DCI} = || F - \bar{\bar{F}} ||^2$.
This ensures that the reconstructed features closely match the original ones, promoting data consistency. 
We use two optimizers: one for training \EA and SSL with $\mathcal{L}_{DT}$, and another for training the \DCI branch with $\mathcal{L}_{DCI}$. This dual-optimizer approach ensures optimal performance for both components.

\noindent\textbf{Sound Source Localization Pipeline.}
\label{downstraem_task}
We implement two DNN-based SSL solutions for Direction of Arrival (DoA) estimation using multi-channel audio: SELDNet~\cite{shimada2021accdoa, adavanne2018sound} and SALSA~\cite{Nguyen_2022}, both utilizing two features (Section~\ref{feature_extraction}).

\textit{SELDNet} is a Convolutional Recurrent Neural Network (CRNN)~\cite{cao2019polyphonic}, combining CNNs for spatial feature extraction and RNNs for temporal dependencies. A fully connected (FC) layer estimates source locations in Cartesian coordinates, providing precise DoA estimates.

\textit{SALSA} is an advanced SSL algorithm with a CRNN model based on the PANN ResNet22 \cite{9229505} architecture. It includes a two-layer Bidirectional GRU and fully connected output layers, adapting to input features by adjusting the number of input channels. During inference, sound classes above the detection threshold are considered active, and their corresponding DOAs are selected.

\subsection{Evaluation Baseline}
\label{baseline}
We evaluate \Sys by comparing it against four imputation methods and one deep interpolation technique, covering a range of strategies from simple imputation to advanced neural network models. This ensures a thorough comparison of \Sys’s effectiveness in handling missing data.

\noindent\textbf{Imputation Methods.} 
We benchmark \Sys against three signal processing and two deep learning imputation methods, including one designed for acoustic signals. Each method is evaluated on both pre-trained and retrained SSL networks, with retraining done using either corrupt or imputed data.

\noindent$\bullet$ \textit{Mean Imputation (\textit{$Mean$}).} Replaces missing channel data with the mean of available channels~\cite{allison2001missing}, a simple and commonly used baseline method.
    
\noindent$\bullet$ \textit{Hot Deck Imputation (\textit{$HotDeck$}).} Fills missing data using the most correlated channels~\cite{little2019statistical}, offering a context-aware alternative to mean imputation.
    
\noindent$\bullet$ \textit{Probabilistic Imputation (\textit{$Prob$}).} Estimates missing data using probabilistic models~\cite{7987674}, handling complexity and randomness in real-world datasets.

\noindent$\bullet$ \textit{Deep Interpolation ($AutoEnc$).} 
Uses autoencoders (AutoEnc) to interpolate missing data, comparing its performance in SSL to evaluate the effectiveness of DNN-based interpolation.

\noindent$\bullet$ \textit{$SoundSeive$.} 
A state-of-the-art hierarchical interpolation method in the STFT domain~\cite{monjur2023soundsieve}, capturing spectral patterns for robust missing data imputation, providing a benchmark for \Sys.

\noindent\textbf{Evaluation Networks.}
We evaluate each method on (1) pretrained and (2) retrained SSL networks.

\noindent$\bullet$ \textit{Pre-trained ($PreTrain$)}. The SSL network (e.g., SELDNet) is trained without missing channel data ($MDP=0\%$) and tested on data with varying $p$\% missing timeframes, assessing performance degradation.

\noindent$\bullet$ \textit{Retrained ($ReTrain$).}
The SSL network is retrained with corrupt or imputed data ($mean$, $HotDeck$, \textit{SoundSeive}, and $AutoEnc$) at different MDP levels, evaluating the robustness of retraining on missing or imputed data.

We evaluate performance using the \textit{Degree of Arrival Estimation Error ($E_{DoA}$)}, which measures the angular difference between the estimated and true sound source locations. $E_{DoA}$ quantifies localization accuracy, providing key insights into system's ability to determine the direction of incoming sounds, critical for spatial audio applications.
\section{Results}
This section begins by evaluating \Sys on various This section evaluates \Sys on various localization algorithms, comparing it to a state-of-the-art method to show current limitations.  For a fair comparison, we compare against both pretrained and retrained versions of all baseline models under the same missing data conditions using the same training and evaluation splits as \Sys. This ensures all models are equally exposed to corrupted input during learning. We analyze performance against baselines under varying missing information and conduct an ablation study to assess \Sys's components. Finally, we test \Sys in a different sensor domain to show its generalizability. Section~\ref{real_world_eval} further evaluates \Sys in real-world environments.


\subsection{Evaluation on Different Sound Source Localization Algorithms}
\begin{figure}
\centering  
\includegraphics[width=.3\textwidth]{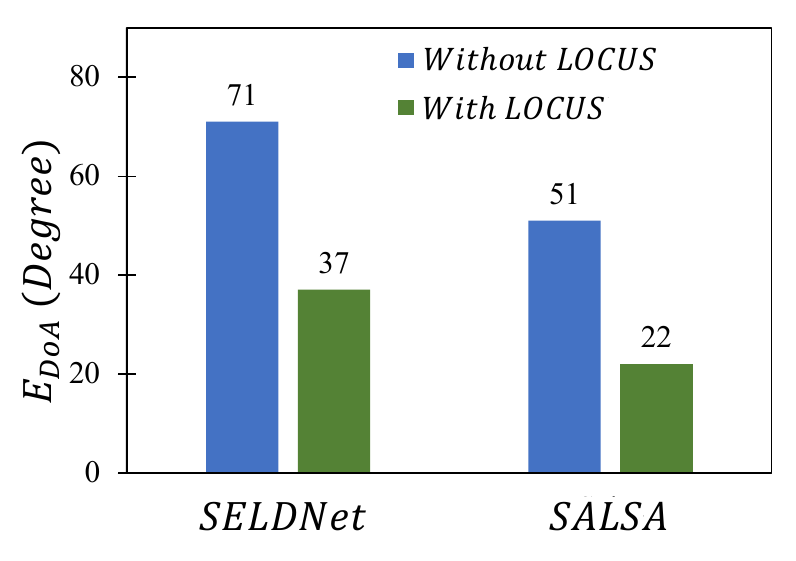}
    \caption{Comparison of \Sys with SALSA at $\textbf{MDP=75\%}$}
    \label{fig:salsa}
\end{figure}

We evaluate \Sys on two sound source localization algorithms -- SELDNet and SALSA. Figure \ref{fig:salsa} illustrates that with no missing information, SALSA and SELDNet achieve $E_{DoA}$ values of 12.74 and 44.7 degrees, respectively. However, when MDP is 75\% or 75\% time-frame are missing channels, the $E_{DoA}$ increases by 38.74 degrees for SALSA and 27.07 degrees for SELDNet. The introduction of \Sys reduces these errors by 10.34 and 7.70 degrees, respectively.

\begin{figure}[t]
\centering
    \includegraphics[width=0.45\textwidth]{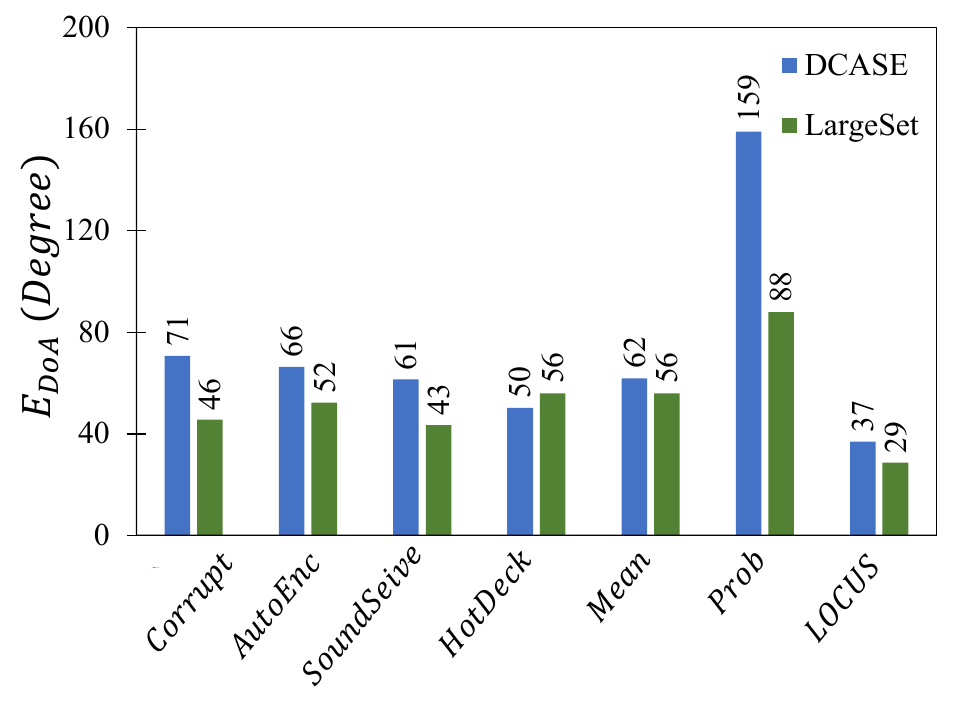}
    \caption{\textbf{Comparison against various imputation methods} evaluated using pre-trained models at MDP = 75\%. \Sys achieves the lowest $E_{DoA}$ across both datasets, while the probabilistic imputation method shows the highest degradation, particularly on DCASE, highlighting the robustness of \Sys under high data loss. }
    \label{comp_dcase_impute}
\end{figure}

\begin{figure}[!htb]
\centering
    \includegraphics[width=0.45\textwidth]{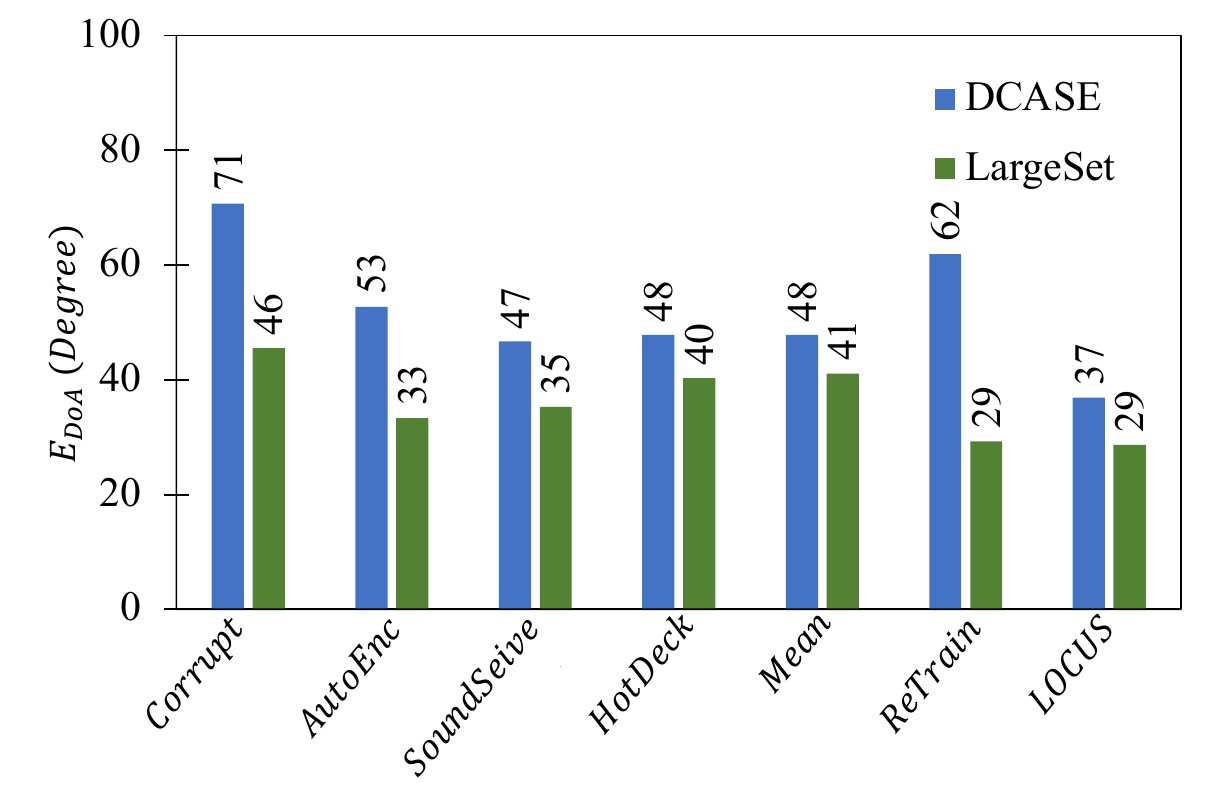}
    \caption{\Sys achieves the lowest $E_{DoA}$ across both datasets compared to \textit{retrained} recovery methods with MDP=75\%, indicating that simple retraining is insufficient to handle high missingness effectively.}
    \label{comp_dcase_dl}
\end{figure}

\begin{figure}[t]
\centering
    \includegraphics[width=0.45\textwidth]{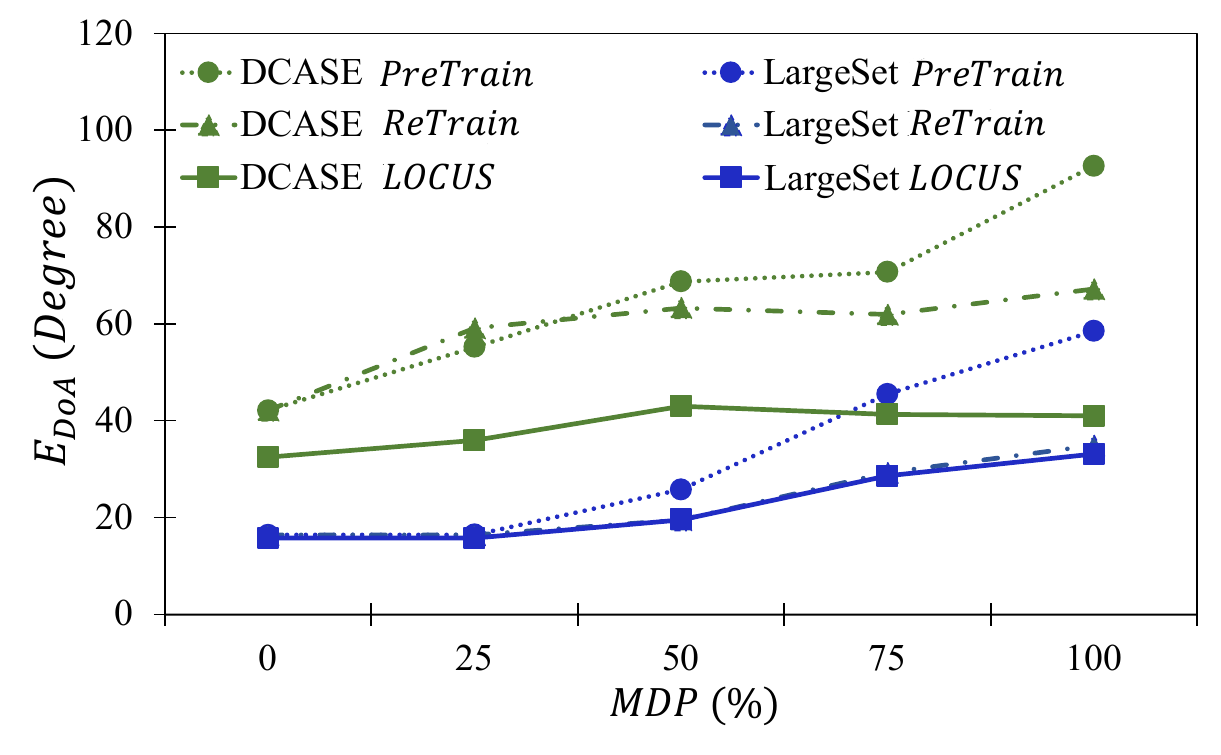}
    \vspace{-1.5em}
    \caption{$E_{DoA}$ versus \textit{MDP\%} for \textit{PreTrain}ed SELDNet, \textit{ReTrain}ed SELDNet, and \Sys with SELDNet across DCASE and LargeSet. \Sys maintains significantly lower error across all MDP levels, highlighting its superior robustness to missing data compared to conventional pretraining or retraining approaches.}
    \label{diff_mdp}
\end{figure}



\begin{table}[]
\caption{Computational complexity and overhead analysis of \Sys, SELDNet, and SALSA. }
\label{tab:complexity_table}
\resizebox{0.49\textwidth}{!}{
\begin{tabular}{@{}lcccc@{}}
\toprule[1.5pt]
Model & \begin{tabular}[c]{@{}c@{}}Number of \\ Parameter\end{tabular} & FLOPS & \begin{tabular}[c]{@{}c@{}}Execution\\ Time\end{tabular} & \begin{tabular}[c]{@{}c@{}}Power \\ Consumption\end{tabular}\\ \midrule[1pt]
SELDNet & 595K & 20B & 1.04s & 120.04J\\ 
\Sys - SELDNet & 3,686K & 28B & 1.08s & 121.64J\\ 
SALSA & 20,313M & 169B & 1.13s & 123.96J\\ 
\Sys - SALSA & 20,560M & 178B & 1.19s & 132.76J\\ \bottomrule[1.5pt]
\end{tabular}
}
\end{table}


Table~\ref{tab:complexity_table} provides a complexity and overhead analysis for SALSA and SELDNet, both with and without \Sys. It highlights that SALSA has $34093.87\times$ more parameters and requires $8.45\times$ more floating-point operations per second (FLOPS) compared to SELDNet. Figure~\ref{fig:salsa} shows that, when combined with \Sys, the simpler SELDNet (with only $595K$ parameters) achieves an $E_{DoA}$ that is $14.01$ degrees lower than SALSA. Moreover, \Sys introduces only a minimal runtime overhead of $3.45 - 5.19\%$ and a slight increase in power consumption of $7.12 - 7.33\%$, which remains lower than SALSA alone.

\subsection{Comparison with Baseline Algorithms}
\label{not_real_world}
We analyze the performance of \Sys against the baseline algorithms, as detailed in Section~\ref{baseline}, using both the DCASE and LargeSet datasets.

\noindent\textbf{Evaluation on Pre-trained SSL Network} 
Figure \ref{comp_dcase_impute} shows that without any imputation the $corrupt$ data suffers from $70.70$ and $45.52$ degree  $E_{DoA}$ on DCASE and LargeSet, respectively, which is $47.89-36.95\%$ more than when all information is present. Three time-domain imputation techniques, $Mean$, $HotDeck$, and $Prob$, reduces this $E_{DoA}$ to $50.14$, $61.91$, and $159.5$ degree for DCASE and $E_{DoA}$ $55.89$, $55.68$, and $87.88$ degree for LargeSet. Imputation performs worse than the $Corrupt$ baseline because it disrupts the inherent time-frequency structure of acoustic signals.

$AutoEnc$ has $66.30$ and $52.30$ degree $E_{DoA}$ for DCASE and LargeSet, respectively which is better than $corrupt$ but still $36.73-57.93\%$ more than when all information is present. The imputation targeted for Acoustic signals, $SoundSieve$ improves performance over both $corrupt$ and $AutoEnc$ by reducing the $E_{DoA}$ to $61.40$ and $43.42$ degree which is still $31.59-49.33\%$ lower than when all information is present.

On the contrary, \Sys, recovers the missing features, and reduces $E_{DoA}$ to $36.90$ and $28.62$ degree. Which is a $13.95 - 76.72\%$ improvement over all the baselines on DCASE. \Sys also outperforms all the compared baselines $32.55- 67.04\%$ on LargeSet.

\noindent\textbf{Evaluation on Retrained SSL Network} 
Figure \ref{comp_dcase_dl} compares all retrained baselines with the proposed \Sys model. Retraining the models with corrupted data improves their overall performance compared to pre-trained models. Although retrained models such as $HotDeck$, $Mean$, and $ReTrain$ reduce $E_{DoA}$, their error rates remain $8.33 - 29.03\%$ higher than when full information is available on DCASE and $29.26 - 24.86\%$ on LargeSet. Further retrained models, $AutoEnc$ and $SoundSeive$, achieve an additional reduction in $E_{DoA}$ by $2.08 - 9.43\%$, yet they still perform $16.98 - 6.38\%$ worse than fully informed DCASE and $37.18 - 33.33\%$ worse on LargeSet. By contrast, \Sys leverages mutual information across multiple channels, recovering performance by $15.19\%$ on DCASE and $24.13\%$ on LargeSet, respectively.

\subsection{Effect of Missing Data Percentage}

Figure ~\ref{diff_mdp} compares the $E_{DoA}$ of \Sys, $PreTrain$, and $ReTrain$ when we have MDP $=$ $25\%, 50\%, 75\%,$ and $100\%$, which is corrupt data time-frames, on DCASE and LargeSet. \Sys achieves $9.6-51.5$ and $0.64-25.4$ degrees lower $E_{DoA}$ than $PreTrain$ on DCASE and LargeSet, respectively. With a higher percentage of missing information, $PreTrain$’s performance degrades significantly, while \Sys consistently retrieves missing information, maintaining variances of $4.39$ and $7.89$ degrees in $E_{DoA}$ for DCASE and LargeSet. Compared to the retrained network with corrupt data ($ReTrain$), \Sys achieves $9.6-26.1$ and $-0.05-1.78$ degrees lower $E_{DoA}$ on DCASE and LargeSet, respectively. Both $PreTrain$ and $ReTrain$ show significant increases in $E_{DoA}$ with higher MDP, as spatial relationships among input streams are crucial for localization tasks. The improvement is less pronounced for LargeSet compared to DCASE because LargeSet is monophonic with minimal MDP impact, whereas the polyphonic DCASE dataset experiences more corruption, highlighting \Sys's performance boost.

\begin{figure}[!htb]
    \centering
    \includegraphics[width=.3\textwidth]{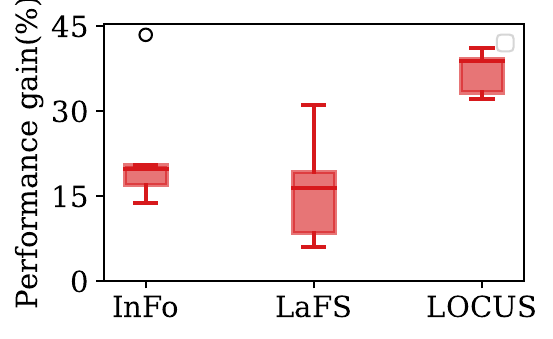}
    \caption{Improvement of $E_{DoA}$ from $RTrain_p$ on DCASE with  $MDP = \{10, 25, 50, 75, 100\}\%$.}
    \label{comp_with_rtrainp}
\end{figure}


\subsection{Ablation Study}
This section evaluates the contribution of each component of \Sys using the DCASE dataset and SELDNet SSL network. \newsn{Our ablation shows that this soft fusion using $I(\tilde{F})$ outperforms alternatives such as direct interpolation or naïve masking.}

\noindent\textbf{Effect of \EA.}
\label{ea_effect}
To understand the impact of \EA, we evaluate \Sys without \DCI, leaving only \EA and the SSL network. Figure~\ref{comp_with_rtrainp} shows that this configuration reduces the $E_{DoA}$ by $22.85\pm 10.57\%$ compared to retrained network with corrupt data. Adding \EA enhances performance by focusing attention on the elements of the feature matrix ($\Tilde{F}$) that contain more information. However, without \DCI to estimate the values of the missing elements in the full-rank version, $F$, this setup performs $14.05 \pm 8.92\%$ worse than the full \Sys implementation.

\noindent\textbf{Effect of \DCI.}
\label{dci_effect}
Next, we investigate the effect of \DCI alone by removing \EA from \Sys. This examination reveals the ability of an interpolator to recover complex missing information and the extent of additional errors introduced during interpolation. Figure~\ref{comp_with_rtrainp} shows that \DCI alone reduces $E_{DoA}$ by $16.22\pm 8.87\%$ compared to retrained network with corrupt data. However, while \DCI estimates the full-rank features ($\bar{\bar{F}}$), it unintentionally introduces errors in elements that are less affected by missing data. The inclusion of \EA helps mitigate this by providing information entropy, which allows the replacement of these polluted components with their original elements, thereby reducing the overall error.

\noindent\textbf{Effect of \GR.}
\GR in \Sys combines the information entropy, $I(\tilde{F})$ from \EA with the interpolated full-rank feature $\bar{\bar{F}}$ from \DCI. As previously discussed, \EA estimates the available information matrix, $\hat{F}$  from ${\tilde{X}}$, while \DCI interpolates to estimate the full-rank feature, $\bar{\bar{F}}$. For feature elements in $\hat{F}_{\tilde{X}}$ significantly impacted by missing channels, \GR replaces them with interpolated elements from $\bar{\bar{F}}$. Conversely, for elements less affected by missing channels, \GR preserves the original features from $\hat{F}_{\tilde{X}}$ using \EA. This method ensures a better estimation of $\hat{F}$, enabling \Sys to outperform both \EA and \DCI alone by $16.22\pm 8.87\%$ and $22.85\pm 10.57\%$, respectively, as shown in Figure~\ref{comp_with_rtrainp}. These observations confirm that retraining the baseline model alone is insufficient and that using \EA or \DCI independently is inadequate. Thus, all three components of \Sys are crucial for recovering corrupted information.

\definecolor{Mercury}{rgb}{0.882,0.882,0.882}
\begin{table}[t]
\small
\centering
\caption{Environment Description for Real World Evaluation. 
}
\label{env_desc}
\begin{tblr}{
  width = \linewidth,
  colspec = {Q[38]|Q[350]|Q[835]},
  row{odd} = {Mercury},
}
\hline[1.5pt]
No  & Dimension ($ft$)             & Description                                                                                                         \\ \hline[1pt]
\#1 & $23.5 \times 46 \times 10$ & A large room with no window and lots of furniture and electrical equipment. \\
\#2 & $18.5 \times 18 \times 10$ & A medium size office space with multiple furniture and large glass windows.                     \\
\#3 & $11 \times 12.5 \times 10$ & A small laboratory with multiple furniture and HVAC    \\ \hline[1.5pt]                                                           
\end{tblr}
\normalsize
\vspace{-1.5em}
\end{table}

\section{Real World Evaluation}
\label{real_world_eval}

This section evaluates \Sys in three real-world environments with two energy scenarios, using ambient energy to power the microphone array.

\noindent\textbf{Energy Harvesting Setup.} 
We use an ETFE solar panel with a step-up regulator and Powercast harvester-transmitter system to harvest RF and solar energy across three environments. The solar panel, placed near a window, experiences intermittence from clouds and shadows, while the RF system faces disruptions from human traffic. Energy traces are collected from multiple MCU locations, each powered by different harvesters.

\begin{figure}[t]
\centering
    \subfloat[Solar\label{solar}]{\includegraphics[width=.48\linewidth]{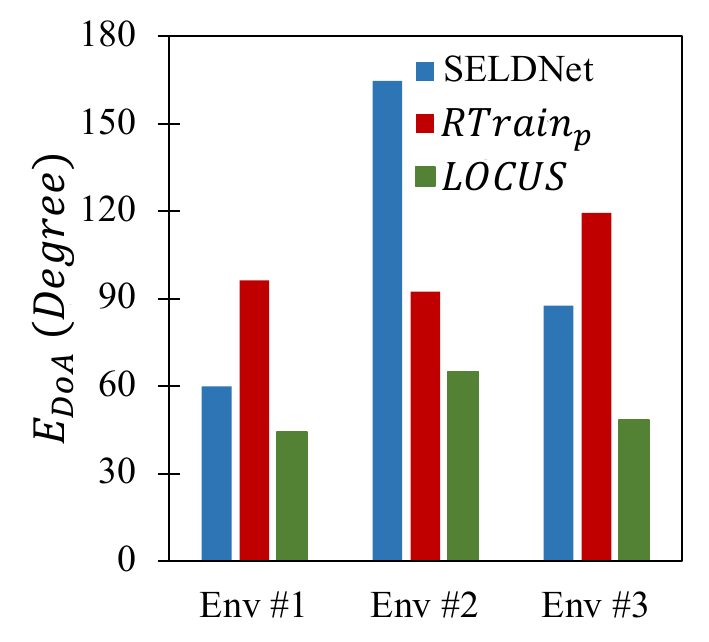}}
    \subfloat[Radio Frequency \label{rf}]{\includegraphics[width=.50\linewidth]{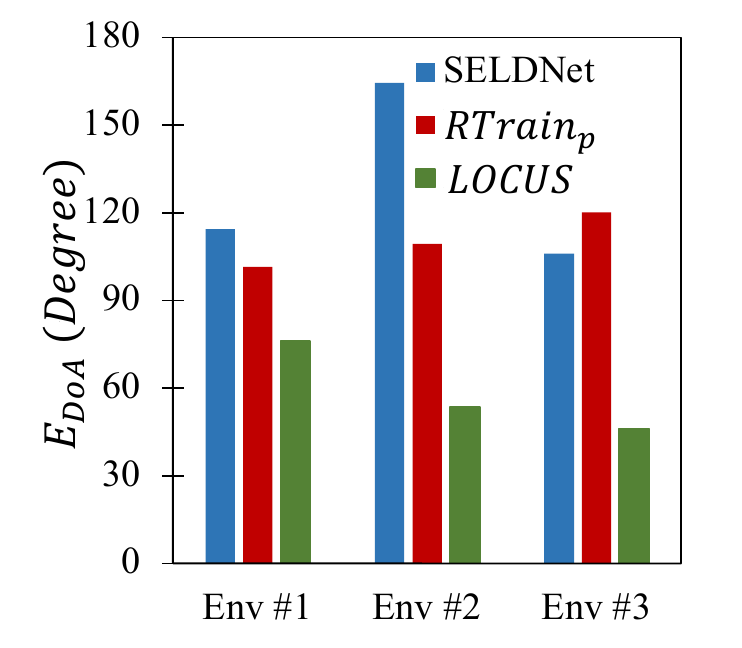}}
    \caption{\Sys with SELDNet recovers missing features more effectively and achieves lower $E_{DoA}$ than $RTrain_{75}$ SELDNet for solar and RF harvested devices.
    }
    \label{real_world_result}
\end{figure}

\noindent\textbf{Acoustic Data Collection Setup.}
We use two MAX78000 microcontrollers to record from a four-channel microphone array, with each MCU connected to two microphones and powered by independent RF or solar harvesters. This distributed setup simulates real-world conditions where environmental factors, such as RF shadowing, multipath, or occlusion, lead to asynchronous power availability across nodes, resulting in partial channel dropouts. Each MCU logs data locally, and synchronization is achieved using periodic sync tones emitted by the sound source and captured by all microphones. This approach allows for offline alignment of data without the need for real-time synchronization. Due to I\textsuperscript{2}S constraints, we employ two MCUs to manage the microphone array, ensuring independent and efficient data collection across the system.

We collect a polyphonic dataset with 1–3 concurrent audio classes, using 96 one-minute segments from seven classes in the NIGENS database~\cite{trowitzsch_2019_2535878}. The speaker is placed at 18 azimuthal angles (10° increments) across three environments (Table~\ref{env_desc}). Precise speaker locations are measured as ground truth, and background noise introduces natural variability. A total of 96 minutes of data are collected, with 72 for training and 24 for evaluation.

\noindent\textbf{Results.}
\label{real_world_eval_section}
Figure~\ref{real_world_result} shows that \Sys outperforms both pre-trained and retrained SELDNet, reducing $E_{DoA}$ by $25.87-60.55\%$ and $29.65-59.46\%$ in solar scenarios. Retrained SELDNet has higher $E_{DoA}$ than the pre-trained model in environments 1 and 3 due to retraining on $MDP=75$\% data, which does not account for stochastic variations. In contrast, \Sys effectively recovers missing information.

In the RF harvester scenario, where energy fluctuations are more frequent, Figure~\ref{rf} shows that RF harvesting leads to more missing channels, causing higher errors for pre-trained SELDNet. Pre-trained and retrained SELDNet show $118-157\%$ and $33-160\%$ higher $E_{DoA}$ than \Sys. In contrast, \Sys only shows $3.84-22.50\%$ more error than the no-missing-channels scenario, demonstrating its robustness.
In contrast to solar harvesting, retrained SELDNet performs better than the pre-trained model in environment 1, suggesting that retraining on a fixed MDP does not adapt well to lower MDP conditions.

\vspace{-1em}
\section{Related Work}

\parlabel{Intermittent System}
Early research in intermittent computing focused on data consistency under unreliable power, using checkpointing strategies to mitigate data loss during power interruptions. Systems like Mementos \cite{10.1145/1950365.1950386} implemented checkpointing but failed to address gaps in data during periods between checkpoints. Recent systems such as Greentooth \cite{10.1145/3649221} and FreeML \cite{farina2024memory} explore resilient architectures for energy-harvesting devices, while SONIC \cite{gobieski2018intermittent} improves DNN inference under intermittent power. Additionally, \cite{majid2020continuous} shows audio sensing on intermittently-powered devices but is limited to short events (~283 ms), insufficient for continuous data collected in real-world applications.

\parlabel{Missing Information Retrieval}
Statistical methods such as mean imputation \cite{allison2001missing}, hot deck \cite{little2019statistical}, and multiple imputations \cite{rubin1996multiple} are used to handle missing data, but these methods are mainly designed for single-channel data and struggle with long sequences of missing data \cite{hasan2021missing}. In the deep learning domain, NLP \cite{katwe2020comparative} and CV \cite{zhang2017learning} have explored missing data recovery, but these approaches mainly focus on text or image restoration. The closest work to our approach is SoundSieve \cite{monjur2023soundsieve}, which addresses missing segments in single-channel audio due to energy intermittence, but it does not consider multi-channel systems. Other speech enhancement works, such as \cite{taherian2022one, mohapatra24_interspeech}, focus on recovering corrupted multi-channel data, but these are not directly applicable to sound source localization, which requires spatial data recovery. Recent multimodal work \cite{ma2022multimodal, ma2021smil} recover missing modalities, but ignores interrelated sensors in a single modality.

\parlabel{Sound Source Localization}
Classic SSL methods, like MUSIC \cite{gupta2015music}, independent component analysis \cite{noohi2013direction}, and sparse models \cite{yang2018sparse}, struggle with under-determined scenarios. Deep learning-based SSL has shown promise \cite{ferguson2018sound, yiwere2017distance}, but fails in presence of missing or corrupted data. 

\parlabel{Attention Mechanism}
Attention mechanisms focus on critical information while ignoring irrelevant data \cite{colombini2014attentional}. In multi-channel input, attention is used to explore channel characteristics and estimate channel state information \cite{gao2021attention}. Although typically applied later in a network, early attention has shown substantial performance improvements \cite{hajavi2020knowing}, motivating the use of \EA in our approach.

\section{Discussions and Limitations}

\begin{figure}[!htb]
\centering
\includegraphics[width=.3\textwidth]{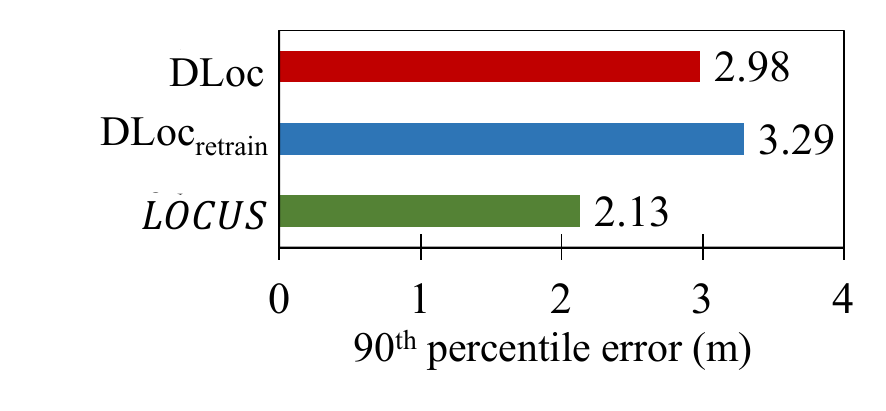}
    \caption{Result of WiFi localization at $MDP=75\%$.
    }
    \label{wifi_loc}
\end{figure}

\parlabel{Beyond Audio and DoA}
\newsn{While \Sys is designed for sound source localization, its feature recovery formulation is modality-agnostic. To demonstrate generalizability, we apply it to Wi-Fi CSI-based localization (\textbf{distance estimation}) using amplitude maps analogous to MFCCs.
We evaluate \Sys on the DLoc model~\cite{ayyalasomayajula2020deep} with a large-scale dataset covering two indoor environments and 16 CSI channels (4 access points × 4 antennas) under 75\% missing data (MDP = 75\%). We adopt DLoc’s two-step architecture: an encoder processes CSI heatmaps, and two decoders predict spatial consistency and location. As shown in Figure~\ref{wifi_loc}, \Sys reduces the 90th percentile localization error to 2.13\,m 35.25\% gain over retrained DLoc (3.29\,m) and 25.52\% over the original (2.98\,m). This demonstrates \Sys’s applicability beyond audio, including in RF sensing with incomplete data.}

\parlabel{Offline Processing}
\Sys is designed for post-hoc analysis, where data is logged during deployment and processed offline, making it suitable for batteryless or energy-constrained systems where real-time inference is impractical. This allows us to employ complex recovery mechanisms without on-device computation costs, though it limits its use in ultra-low-latency applications. Future work could explore lightweight, on-device approximations and integrate recovery-aware sensing strategies. Additionally, while we use sync tones for alignment, drift-aware synchronization could be explored for long-term distributed deployments.

\begin{figure}[t]
    \centering
    \vspace{-1em} 
    \includegraphics[width=0.38\textwidth]{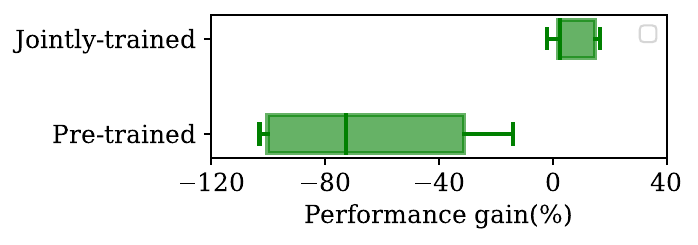}
    \vspace{-1em}
    \caption{Impact of joint training on $E_{DoA}$.}
    \label{training_effect}
    \vspace{-1em}
\end{figure}

\parlabel{Impact of Joint Training}
\Sys is jointly trained with the localization network to address feature distribution shifts caused by \EA and \DCI. As shown in Figure~\ref{training_effect}, joint training improves performance significantly. The recovery blocks are modular and can be trained separately for integration into existing systems, with future work exploring generalization strategies to reduce retraining requirements.

\parlabel{Generative Models for Information Retrieval}
Generative models like VAEs and GANs are powerful but may struggle with reconstruction accuracy, particularly for discrete data. While they could replace the auto-encoder in \DCI, we find that a simpler auto-encoder performs comparably with less training overhead, making it a more efficient choice for multi-channel information retrieval in \Sys.

\parlabel{\Sys for Multi-Sensor Systems}
While this paper focuses on multi-sensor systems like microphone arrays, \Sys could be extended to multimodal systems (e.g., video, audio, text). For instance, \Sys could recover missing audio data from video or text sources in a video conferencing setup, improving the overall quality.

\parlabel{Microphone Geometry Dependency}
\Sys relies on microphone-array geometry for localization \newsn{and functions as a preprocessing layer that reconstructs reliable features before they are passed to downstream SSL architectures. This design allows us to support existing geometry-dependent methods without requiring modifications to their core structure.} Future work will explore domain generalization techniques to address this limitation.

\parlabel{Large Real-World Localization Dataset}
We created and released a 50-hour simulated dataset to train DNNs for sound source localization. However, simulated data may lack the complexity of real-world environments, and future work will focus on developing a comprehensive real-world dataset for sound source localization.

\parlabel{MDP incorporating number of missing channels} \newsn{While incorporating the number of missing channels could add granularity, we excluded it since MDP is solely an evaluation metric, not used during training. Using a consistent MDP across varying missing-channel scenarios introduces a more challenging evaluation, better showcasing our method’s robustness; jointly modeling temporal extent and channel count remains a promising direction for future work.}

\parlabel{Image Recovery Limitations for Acoustic Features}
While image pixel recovery methods are effective for images, they are unsuitable for frequency-domain features like GCC and MFCC, which do not exhibit the same neighbor similarity. Thus, we cannot directly apply image recovery methods to restore missing acoustic channel data.

\parlabel{Handling Sparse Inputs}
While \Sys is not designed to operate on sparse inputs, it reconstructs missing features to restore dense representations before localization. Future work could explore architectures operating on sparse or partially observed features to reduce reconstruction overhead.

\parlabel{System-Level Exploration}
A full exploration of communication overhead, runtime efficiency, and hardware acceleration is beyond the scope of this work. However, future efforts could focus on lightweight on-device approximations and energy-aware communication strategies for embedding \Sys in fully deployed systems. \newsn{To support on-device adaptation, we plan to explore a more comprehensive set of baselines and adopt efficient model architectures tailored for resource-constrained environments. Additionally, we plan to reduce feature dimensionality and apply model compression techniques such as pruning and quantization \cite{biswas2025quads} to minimize memory and computational demands.
}

In this paper, we present \Sys, an information retrieval-based sound source localization system designed to perform effectively despite missing or corrupted data. By leveraging the interdependence of data across channels, \Sys recovers missing information that traditional algorithms, which require prior knowledge and struggle with high-complexity data, cannot handle. The \EA component quantifies both \emph{how} much and \emph{where} the data is corrupted, while \DCI interpolates the missing features. Our experiments demonstrate that \Sys achieves up to 36.91\% lower degree of arrival (DoA) error compared to retrained models using missing-channel data in the DCASE dataset. Additionally, in real-world scenarios with stochastically missing information, \Sys shows performance gains of 25.87-59.46\%, highlighting its robustness and effectiveness in real-world conditions.
\section*{Acknowledgment}
This research was supported by funding from the NSF CNS-2347692. We gratefully acknowledge their support in enabling this work.
\bibliographystyle{ACM-Reference-Format}
\bibliography{ref}


\end{document}